%% file: main.tex
\documentclass[letterpaper, 10 pt, conference]{ieeeconf}  

\IEEEoverridecommandlockouts                              

\usepackage[shortcuts,acronym]{glossaries}
\usepackage{url}
\newacronym{cnn}{CNN}{Convolutional Neural Network}
\newacronym{scm}{SCM}{Sparse Convolution Module}
\newacronym{ah}{AH}{Adaptive Height}
\newacronym{fh}{FH}{Fixed Height}
\newacronym{rcs}{RCS}{Radar Cross Section}
\newacronym{cagf}{CaGF}{Confidence-aware Gated Fusion}
\newacronym{fmcw}{FMCW}{Frequency-Modulated Continuous-Wave}
\newacronym{cafnet}{CaFNet}{Confidence-aware Fusion Net}

\usepackage{times}
\usepackage{epsfig}
\usepackage{graphicx}
\newsavebox\mybox
\usepackage{amsmath}
\usepackage{amssymb}
\usepackage{subcaption}
\usepackage{amsmath}
\usepackage{float}
\usepackage{xspace}
\usepackage[compress]{cite}
\usepackage[flushleft]{threeparttable}
\usepackage[dvipsnames]{xcolor}
\usepackage{amssymb} 
\usepackage{multirow} 
\usepackage{multicol} 
\definecolor{myyellow}{RGB}{232,216,129}
\definecolor{myred}{RGB}{166,78,60}
\usepackage[font=small,labelfont=bf]{caption}

\begin{document}
\title{\LARGE \bf
CaFNet: A Confidence-Driven Framework for
Radar Camera\\ Depth Estimation
}

\author{\authorblockN{{Huawei Sun$^{1,2}$, Hao Feng$^{2}$, Julius Ott$^{1,2}$, Lorenzo Servadei$^{2}$, Robert Wille$^{2}$ }}\\
\authorblockA{$^{1}$Infineon Technologies AG, Neubiberg, Germany\\
$^{2}$Technical University of Munich, Munich, Germany\\
E-mail: \{huawei.sun, julius.ott\}@infineon.com\\
\{hao.feng, lorenzo.servadei, robert.wille\}@tum.de}}



\maketitle
\thispagestyle{empty}
\pagestyle{empty}

\begin{abstract}
Depth estimation is critical in autonomous driving for interpreting 3D scenes accurately. Recently, radar-camera depth estimation has become of sufficient interest due to the robustness and low-cost properties of radar. Thus, this paper introduces a two-stage, end-to-end trainable \acrfull{cafnet} for dense depth estimation, combining RGB imagery with sparse and noisy radar point cloud data. The first stage addresses radar-specific challenges, such as ambiguous elevation and noisy measurements, by predicting a radar confidence map and a preliminary coarse depth map. A novel approach is presented for generating the ground truth for the confidence map, which involves associating each radar point with its corresponding object to identify potential projection surfaces. These maps, together with the initial radar input, are processed by a second encoder. For the final depth estimation, we innovate a confidence-aware gated fusion mechanism to integrate radar and image features effectively, thereby enhancing the reliability of the depth map by filtering out radar noise. Our methodology, evaluated on the nuScenes dataset, demonstrates superior performance, improving upon the current leading model by $3.2\%$ in Mean Absolute Error (MAE) and $2.7\%$ in Root Mean Square Error (RMSE). Code: \url{https://github.com/harborsarah/CaFNet}
\end{abstract}

\section{Introduction}
\input{chapter/intro.tex}
\label{sec:intro}

\section{Related Work}
\input{chapter/related_work.tex}
\label{sec:related_work}

\section{Approach}
\input{chapter/approach.tex}
\label{sec:approach}


\section{Experiments}
\input{chapter/experiments.tex}
\label{sec:experiments}


\section{Conclusion}
\input{chapter/conclusion.tex}
\label{sec:conclusion}

\section{Acknowledgement}
\input{chapter/acknow}
\label{sec:acknowledgement}








\bibliographystyle{IEEEtran}
\bibliography{references}

\end{document}

%% file: chapter/intro.tex
Dense depth estimation is an important task in autonomous driving, which helps to understand the 3D geometry of outdoor scenes. 
Following the first learning-based monocular depth estimation model \cite{saxena2005learning}, there have been studies enhancing task performance based on deep learning techniques~\cite{cspn,bts,adabins,geonet, yuan2022neural,hu2019revisiting}. Despite these advancements, the efficacy of monocular image-based methods is inherently constrained by the lack of depth cues in RGB images. To address this, numerous studies \cite{sparse-to-dense,nonlocal_spn,joint3d} have incorporated lidar data alongside images for depth completion tasks, yielding higher-quality depth maps.

While integrating lidar data offers improved scene understanding, lidar sensors are known to be sensitive to lighting and weather conditions. In contrast, radar sensors, 
present a cost-effective and weather-adaptable alternative. Moreover, radar sensors capture the velocity of moving objects, yielding more accurate object detection \cite{crfnet,mcafnet,centerfusion}. With the release of autonomous driving datasets \cite{nuscenes,fog} that include radar sensors, the domain of camera-radar fusion is gaining traction. However, radar point clouds are notably sparser compared to lidar point clouds. Additionally, commonly used radar sensors often lack elevation resolution due to limited antennas along the elevation axis, leading to the absence of height information in radar points. 
Moreover, the phenomenon of the radar multipath effect further complicates this scenario by introducing numerous ghost targets.

\begin{figure}[ht]
\centering
\vspace{-2mm}
\includegraphics[width=0.98\linewidth]{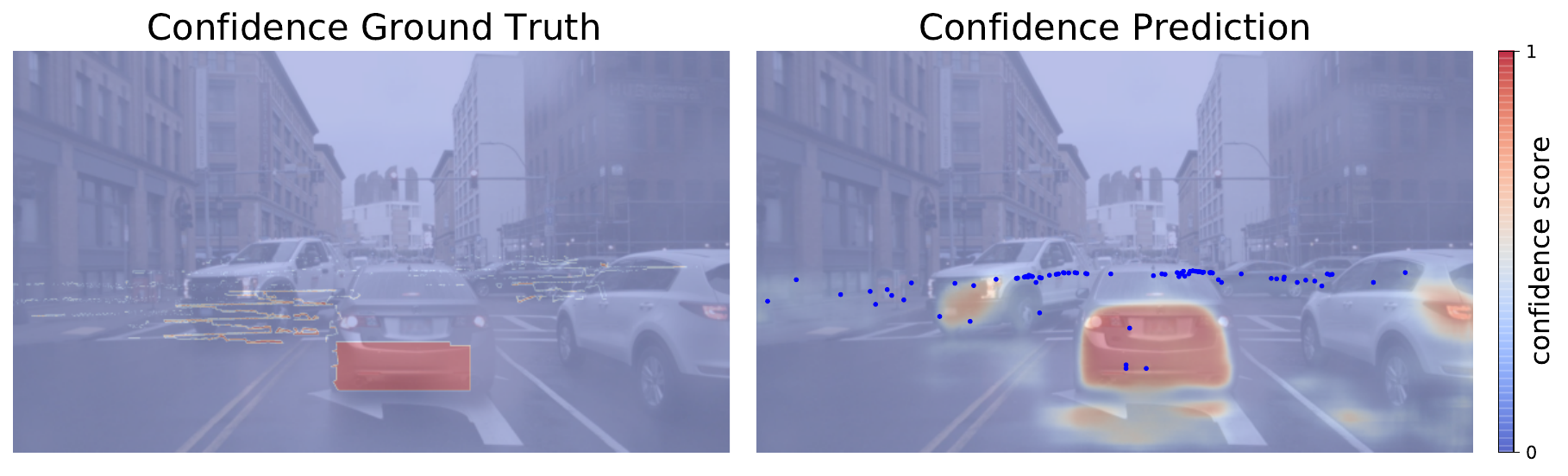}

\vspace{-2mm}
   \caption{Radar confidence map comparison. The ground truth is generated by our method, comparing the depth value of the radar point with the ground truth depth map within a selective region.}
\label{fig:confidence}
\vspace{-2mm}

\end{figure}

\begin{figure*}[ht]
\vspace{0.18cm}
\centering
\includegraphics[width=0.98\textwidth]{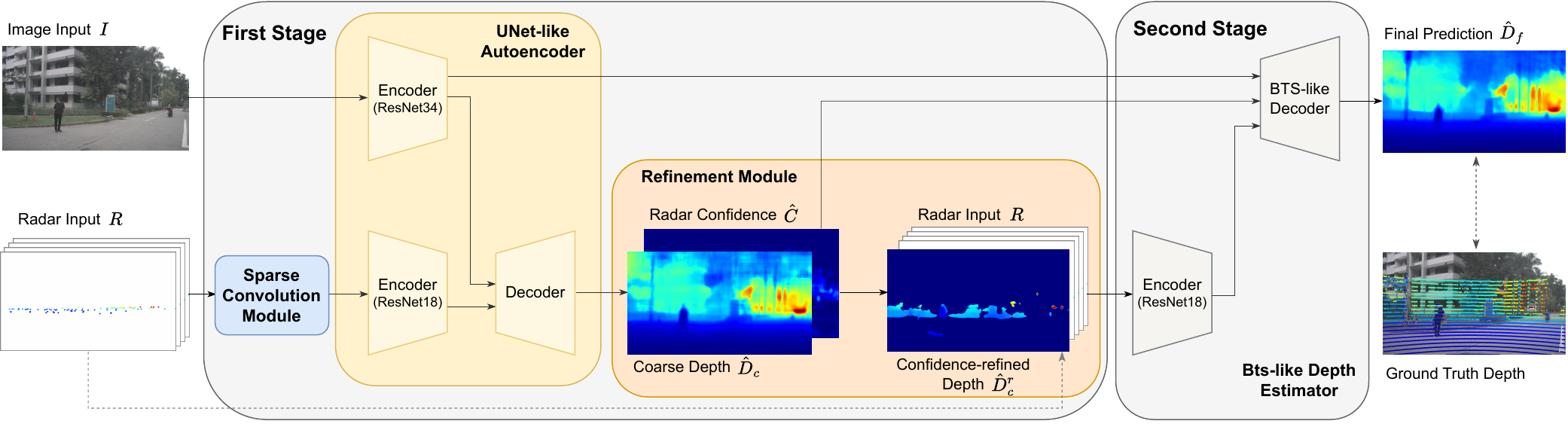}

   \caption{\acrshort{cafnet} model architecture, including two end-to-end trainable phases. Initially, the first stage focuses on estimating a radar confidence map and a coarse depth map. Subsequently, through the refinement module, a confidence-enhanced depth map is merged with the original radar input and then forwarded to a secondary radar encoder. The extracted radar features and image features, which, alongside the predicted confidence map, are input into the final decoder to generate the dense depth map.}
\label{fig:model}
\vspace{-4mm}
\end{figure*}
Given these challenges, effectively leveraging radar points while mitigating noise is a crucial concern. Several studies \cite{dorn_radar,crfnet,mcafnet,enhance} attempt to solve the sparsity and ambiguous height problems by extending radar points along the elevation axis, bringing additional errors into the radar data.
In an alternative approach, RC-PDA \cite{rc-pda} and RadarNet \cite{radarnet} employ two-stage networks that initially estimate radar confidence scores to generate semi-dense depth maps for further training. However, a predefined region with $h\times w$ is selected for each radar point. This treats each radar point uniformly and does not consider the association between the radar point and its corresponding object in the 3D space. 

To overcome the obstacles, we propose \acrshort{cafnet}, a novel two-stage, end-to-end trainable network for depth estimation.
Our method incorporates a \acrfull{scm}, employing sparse convolutional layers \cite{sparsecnn} to handle radar data sparsity before being processed by the encoder. 
In the first stage, a radar confidence map along with a coarse depth map is estimated, which guides the training of the second stage. To better guide the confidence map, we propose a ground truth generation method that considers the association between the radar points and the detected objects.
An example of the confidence maps is illustrated in Fig. \ref{fig:confidence}. 
The key innovation in our work is introducing \acrfull{cagf} in the second stage. This fusion technique integrates radar and image features by taking into account the confidence scores of individual radar pixels.
Our model leverages a single radar scan and a single RGB image for depth estimation. The \acrshort{cafnet} outperforms existing solutions, establishing new benchmarks on the nuScenes dataset \cite{nuscenes}.

To summarize, our contributions are as follows:
\begin{itemize}
    \item We introduce the \acrfull{cafnet}, which surpasses previous state-of-the-art methods in radar-camera depth estimation on the nuScenes dataset.
    \item A novel approach for generating the ground truth of the radar confidence map is presented, enhancing the reliability of guiding the confidence generation.
    \item The \acrfull{cagf} technique is proposed, effectively reducing the propagation of erroneous data, thereby improving the overall performance
    of the depth estimation.
\end{itemize}

%% file: chapter/related_work.tex
\subsection{Monocular Depth Estimation}
Monocular depth estimation is the process of determining the depth of each pixel in an image using a single RGB image. 
The majority of current methods employ encoder-decoder architectures to derive depth maps from input images. A significant advancement in this field was made by Eigen \textit{et al.} \cite{eigen2014depth}, who introduced the use of Convolutional Neural Networks (CNNs) in a multi-scale network for depth estimation. This included the development of a scale-invariant loss function to address the issue of scale ambiguity.
DORN \cite{dorn} first reinterpreted depth regression as a classification task, focusing on predicting depth ranges instead of exact values with predefined bin centers. Subsequent studies, such as AdaBins \cite{adabins} and BinsFormer \cite{binsformer}, enhanced this approach by learning to predict more accurate bin centers, addressing the variation in depth distribution across different frames.
Incorporating geometric priors has also been shown to be effective. For instance, GeoNet \cite{geonet} uses a geometric network to simultaneously infer surface normals and depth maps that are geometrically consistent. As proposed in BTS \cite{bts}, the local planar assumption guides the upsampling modules in the decoding phase. P3Depth \cite{p3depth} introduces an approach that selectively leverages information from coplanar pixels, using offset vector fields and a mean plane loss to refine depth predictions.
Another line of research focuses on refining coarse depth maps through spatial propagation networks, initially proposed in SPN \cite{spn}. CSPN \cite{cspn} improved this by using convolutional layers to learn local affinities, which better connect each pixel with its neighboring pixels.

\subsection{Camera and Radar-based Depth Estimation}
Camera radar fusion-based techniques combine sparse radar point clouds with camera images for depth map prediction. These methods present more challenges than lidar-camera fusion due to the inherent sparsity and noise in radar point clouds.
Lin \textit{et al.} \cite{lin2020depth} developed a two-stage encoder-decoder network specifically designed to filter out noisy radar points, using lidar points as a reference. An extension of this approach by \cite{dorn_radar} involves augmenting each radar point's height value to produce a denser radar projection, which is then integrated with camera images using a regression network based on DORN's \cite{dorn} methodology.
RC-PDA \cite{rc-pda} tackles the uncertainty of projecting radar points onto the image plane by first learning a radar-to-pixel association. Once the association is established, a second network estimates depth.  
These studies typically use multiple radar sweeps to enhance the density of the point cloud. Singh \textit{et al.} \cite{radarnet} propose a two-stage network, where the first stage learns a one-to-many mapping, generating quasi-dense radar depth maps, which are then processed in the second stage for final depth estimation. 
However, a significant drawback of the two-stage networks in RC-PDA \cite{rc-pda} and RadarNet \cite{radarnet} is their separate training phases. This necessitates local storage of intermediate data post the initial training stage, resulting in a time-intensive process that challenges real-time prediction. Moreover, a predefined number of radar points are selected in RadarNet \cite{radarnet}, with the empirical evidence suggesting a mere four points per frame. This severely constrains the utility of radar data. 

%% file: chapter/approach.tex
This section introduces the principal innovations of our work. Initially, we detail the comprehensive architecture of our \acrshort{cafnet} in Sec. \ref{subsec:model}. Subsequently, Sec. \ref{subsec:radar_conf} describes our novel approach for generating the ground truth of radar confidence maps. Following this, the refinement module is described in Sec. \ref{subsec:refinement_module}. Finally, we present the decoder, incorporating our newly proposed \acrshort{cagf} technique.

\subsection{Model Architecture}
\label{subsec:model}
As illustrated in Fig. \ref{fig:model}, our \acrshort{cafnet} processes an RGB image $I\in \mathbb{R}^{H\times W\times 3}$ and a radar-projected image $R\in \mathbb{R}^{H\times W\times C_{R}}$ as inputs. 
Typically, the ground truth depth $D_{gt}\in \mathbb{R}^{H\times W}$ generated by lidar is sparse. Thus, for the supervision, following \cite{radarnet}, we accumulate point clouds by projecting from neighboring frames to form a denser ground truth map $D_{acc}\in \mathbb{R}^{H\times W}$.
The radar-projected image is generated by directly projecting every radar point onto the image plane, carrying specific information, for example, velocities and \acrfull{rcs}.
In the first stage, feature extraction from the RGB image is conducted using a ResNet-34 \cite{resnet} backbone. In contrast, the radar input undergoes initial refinement through the \acrshort{scm}, which stacks four sparse convolution layers \cite{sparsecnn}, before being encoded by a ResNet-18 backbone. These multi-scale image and radar features are then concatenated and fed into a UNet decoder \cite{unet}, which leverages skip connections to efficiently transmit information from the encoder to the decoder. The decoder outputs two maps. The first output is a coarse depth map $\hat{D}_{c}\in \mathbb{R}^{H\times W}$, which is trained against the ground truth depth map $D_{acc}\in \mathbb{R}^{H\times W}$. The second is a radar confidence map $\hat{C}\in [0,1]^{H\times W}$, which signifies the probable correspondence between projected radar points and image pixels and is trained against a binary ground truth confidence map $C\in \{0,1\}^{H\times W}$, detailed in Sec. \ref{subsec:radar_conf}.

The refinement module, utilizing $\hat{C}$ and $\hat{D_{c}}$ as inputs, produces a confidence-refined depth map $\hat{D}_{c}^{r}\in \mathbb{R}^{H\times W}$. This map is concatenated with $R$, yielding a new radar input $R'\in \mathbb{R}^{H\times W\times (C_{R}+1)}$ and forwarded to the second stage. Sec. \ref{subsec:refinement_module} provides a more detailed explanation.

During the second stage, $R'$ is encoded via a separate ResNet-18 backbone. Subsequently, a BTS-like depth estimator \cite{bts} is employed, which takes the newly extracted radar features, the image features, and the predicted radar confidence as inputs. Throughout this stage, features from different layers are fused using a \acrshort{cagf} mechanism. Leveraging multi-scale features, a final depth map $\hat{D}_{f}\in \mathbb{R}^{H\times W}$ is estimated. Further details can be found in Sec. \ref{subsec:fusion}.


\subsection{Radar Confidence Map}
\label{subsec:radar_conf}
In the first stage in \cite{radarnet}, a set of $K$ radar points is chosen as input. For each point, $K$ ground truth confidence maps of size $288 \times 900$ are generated. However, with its substantially large predefined area, this approach results in significant computational demands. Moreover, since their model cannot handle various numbers of points as input, they select only four radar points during training, which leads to inefficiency.

In contrast, our method diverges by incorporating all radar points in each frame. We also pay particular attention to the relationship between radar points and the associated objects to more accurately determine probable radar projection areas. Specifically, for a frame, we work with a radar point cloud comprising $M$ points, $\mathbf{z}=\{z_{m}\}_{m=1}^{M}$, and an accumulated ground truth depth map, $D_{acc}$. Additionally, we identify $\mathcal{B}=\{B^{1}, B^{2}, \ldots, B^{n}\}$ as a set of object bounding boxes within the frame, where $n$ is the number of detected objects.

\begin{figure}[htbp!]
\vspace{-1mm}
\centering
     \includegraphics[width = 0.92\linewidth]{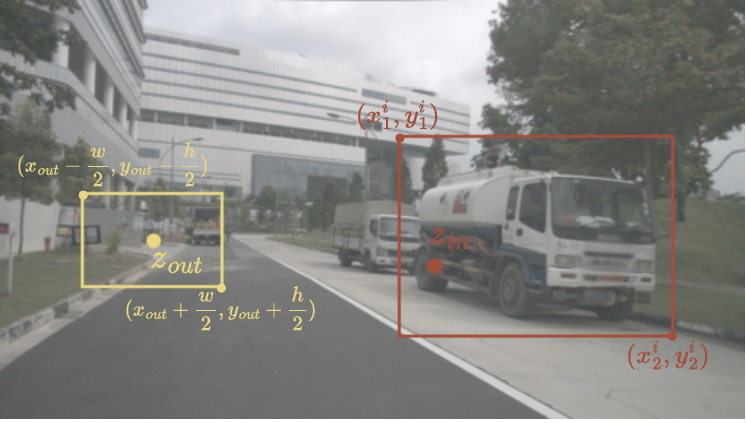}
     \caption{Radar confidence selective region. \textcolor{myred}{red: }radar point is located in an object, \textcolor{myyellow}{yellow: }radar point is not associated with any object.}
     \label{Fig:selective region}
     \vspace{-3mm}
 \end{figure}

We categorize the radar points into two groups. The first, $\mathbf{z_{in}}=\{z_{in,p}\}_{p=1}^{P}$, includes points inside any 3D bounding box, where $z_{in} \in B^{i}$ signifies that the point is in the $i^{th}$ box. The second group, $\mathbf{z_{out}}=\{z_{out,q}\}_{q=1}^{Q}$, contains points not associated with any objects, typically regarded as noise. Fig. \ref{Fig:selective region} depicts these point types and their specific regions.

For a point $z_{in}$ at pixel coordinates $(x_{in},y_{in})$ linked to bounding box $B^{i}$, the selected region $\mathcal{R}_{in}$ is defined as:
\begin{equation}
\mathcal{R}_{in}= \{(x,y) \,|\, x_{1}^{i} \leq x \leq x_{2}^{i}, \, y_{1}^{i} \leq y \leq y_{2}^{i}\},
\end{equation}
where $(x_{1}^{i}, y_{1}^{i})$ and $(x_{2}^{i}, y_{2}^{i})$ represent the top-left and bottom-right corners of $B^{i}$, respectively.

Conversely, for a point $z_{out}$ at coordinates $(x_{out},y_{out})$, the region $\mathcal{R}_{out}$ is selected using a pre-defined patch size $(w, h)$:
\begin{equation}
\vspace{-1.5mm}
\centering
\begin{aligned}
       & \begin{split}
       \mathcal{R}_{out}= \{(x,y)| & x_{out}-\frac{w}{2}\leq x \leq x_{out}+\frac{w}{2}, \\
                                   & y_{out}-\frac{h}{2}\leq y \leq y_{out}+\frac{h}{2}\}.
       \end{split}
    \end{aligned}
\end{equation}

The ground truth confidence is generated by comparing the absolute depth difference of the point's depth with $D_{acc}$ within the selective region. For the $m^{th}$ radar point $z_{m}$ with depth $d_{m}$, the confidence value of each pixel within its selective region $\mathcal{R}_{m}$:
\begin{equation}
    C(i,j) = \left\{ 
  \begin{array}{ c l }
    1 & \quad \textrm{if } |D_{acc}(i,j)-d_{m}|\leq \tau \\
    0                 & \quad \textrm{otherwise.}
  \end{array}
\right.
\end{equation}
Here, $D_{acc}(i,j)$ represents the ground truth depth of pixel $(i,j)$  and $\tau$ is the tolerance threshold, chosen as 0.4 meters.

\subsection{Refinement Module}
\label{subsec:refinement_module}
To facilitate the learning of information from $\hat{D}_{c}$ and $\hat{C}$ by the second radar encoder, our initial concept involved concatenating these two maps with $R$. However, due to the sparsity property of $R$, the detailed information from $\hat{D}_{c}$ might overwhelm the encoder with excessive information. To address this, we introduce a refinement module designed to streamline the information content from $\hat{D}_{c}$, utilizing the predicted radar confidence map as guidance.

In this refinement module, we generate a radar confidence mask $\hat{M}\in \mathbb{R}^{H\times W}$ by setting a predefined threshold $T$:
\begin{equation}
   \hat{M}(i,j) = \left\{ 
  \begin{array}{ c l }
    1 & \quad \textrm{if } \hat{C}(i,j)\geq T \\
    0                 & \quad \textrm{otherwise.}
  \end{array}
\right.
\end{equation}
Followed by this, the confidence-refined depth $\hat{D}_{c}^{r}$ is calculated by element-wise multiplication between $\hat{M}$ and $\hat{D}_{c}$
\begin{equation}
    \hat{D}_{c}^{r} = \hat{M} \odot \hat{D}_{c}.
\end{equation}
After the refinement process, $\hat{D}_{c}^{r}$ is concatenated with $R$, creating a new input $R'$ and then fed into the second stage.
\begin{figure}[ht]
\vspace{-2mm}
\centering
     \includegraphics[width = 0.90\linewidth]{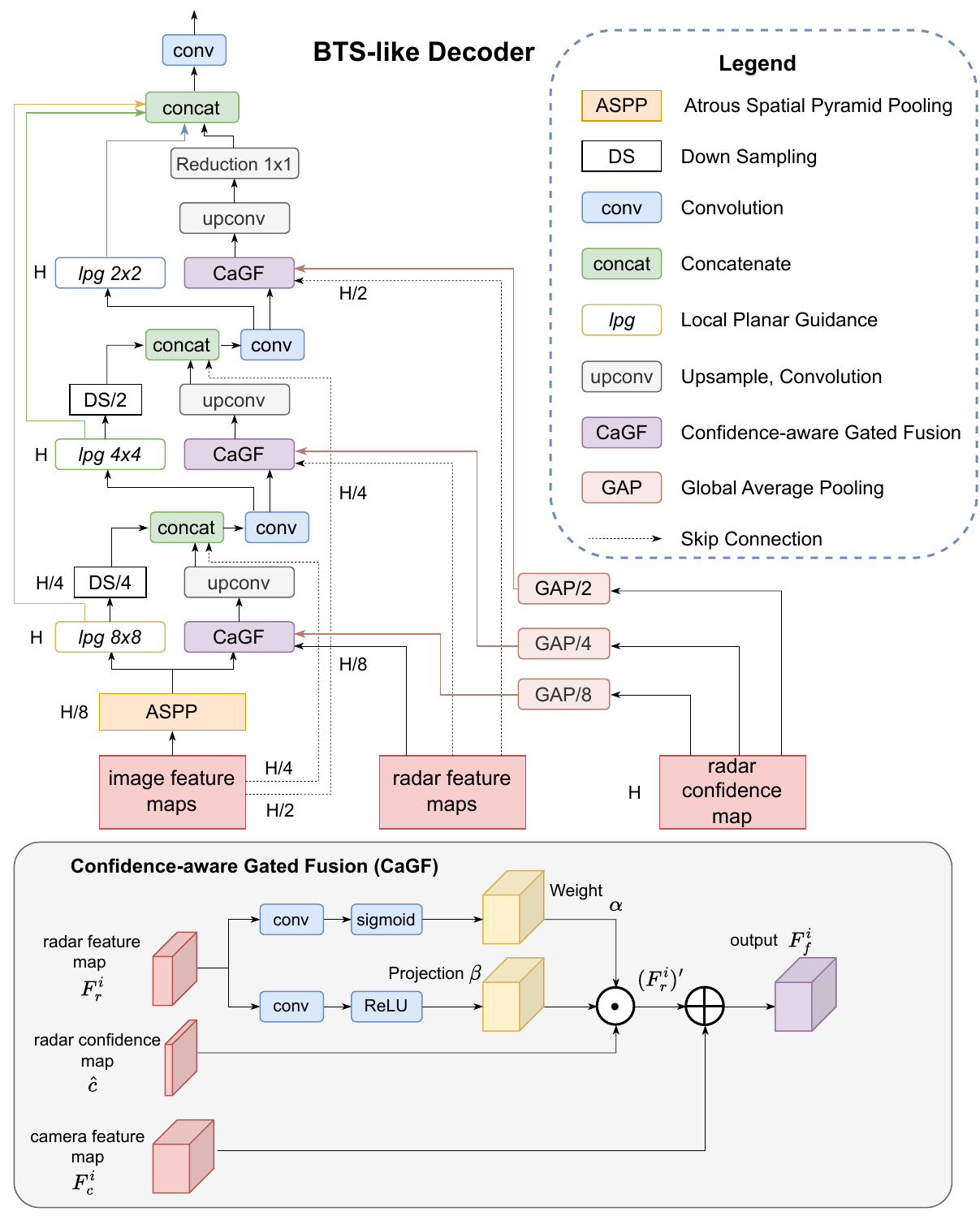}
     \caption{BTS-like Decoder.}
     \label{Fig:bts}
     \vspace{-5mm}
 \end{figure}
\subsection{Bts-like Depth Estimator}
\label{subsec:fusion}
Our final depth estimation model adopts the framework of BTS \cite{bts}. To enhance its performance, we introduce the \acrshort{cagf}—a novel mechanism designed to integrate image and radar features. This process incorporates $\hat{C}$, enabling the selective filtration of noisy radar data. The decoder's architecture is depicted in Fig. \ref{Fig:bts}.

\noindent \textbf{Confidence-aware Gated Fusion: }At the $i^{th}$ stage, the radar feature $F_{r}^{i}$ and the image feature $F_{c}^{i}$ are in shape of $\frac{H}{2^{i}}\times \frac{W}{2^{i}}\times C_{r}$ and $\frac{H}{2^{i}}\times \frac{W}{2^{i}}\times C_{c}$, respectively. Building upon the approach in \cite{radarnet}, we derive a set of weights $\alpha=\sigma(p^{T}F_{r}^{i}) \in [0,1]$ and projections $\beta=q^{T}F_{r}^{i}$ from $F_{r}^{i}$. Here, $p$ and $q$ denote trainable parameters, and $\sigma$ symbolizes the sigmoid activation function, ensuring $\alpha$ scales between 0 and 1. Additionally, we process the predicted radar confidence $\hat{C} \in [0, 1]^{H\times W}$ to obtain a resized confidence $\hat{c}$ by employing Global Average Pooling (GAP) with a stride of $2^{i}$ on $\hat{C}$.

The enhancement of the radar feature is formulated as $(F_{r}^{i})'=\alpha \cdot \beta \cdot \hat{c}$. This refinement step leverages the radar confidence to more effectively distinguish the significance of various positions, a critical consideration given that the majority of radar input pixels are typically zero-valued. Consequently, the composite feature for this stage is obtained as $F_{f}^{i}=(F_{r}^{i})' + F_{c}^{i}$, integrating both radar and image features for a more nuanced representation.

\begin{table*}[ht]
\caption{Performance Comparison on nuScenes Official Test Set.}
\vspace{-1mm}
\label{tab:exp}
\resizebox{17.5cm}{!} 
{
\centering

\begin{tabular}{c||c|c|c|cccccccc}
\hline
\multirow{2}{*}{Max Eval Distance} & \multirow{2}{*}{Method} & \multicolumn{2}{c|}{Sensors} & \multicolumn{8}{c}{Metrics} \\ \cline{3-12} 
                                    &                         & Image      & Radar      & MAE $\downarrow$ & RMSE $\downarrow$ & AbsRel $\downarrow$  &log10 $\downarrow$ & RMSElog $\downarrow$ & $\delta_{1}$ $\uparrow$& $\delta_{2}$ $\uparrow$ & $\delta_{3}$ $\uparrow$  \\ \hline
\multirow{6}{*}{50m}                & BTS \cite{bts}  & \checkmark &            & 1.937 & 3.885 & 0.116                                            &0.045 & 0.179 & 0.883 & 0.957 & 1.937 \\ 
                                    & RC-PDA \cite{rc-pda}   & \checkmark & \checkmark & 2.225 & 4.159 &  0.106   & 0.051  & 0.186 & 0.864 & 0.944 & 0.974      \\ 
                                    & RC-PDA-HG \cite{rc-pda}  & \checkmark & \checkmark & 2.210 & 4.234 & 0.121   & 0.052  & 0.194  & 0.850  & 0.942  & 0.975       \\ 
                                    & DORN  \cite{dorn_radar}  & \checkmark & \checkmark & 1.898 & 3.928 & 0.100 &   0.050 & \textbf{0.164} & 0.905 & 0.962 & 0.982  \\ 
                                    & RadarNet   \cite{radarnet}  & \checkmark &  \checkmark & 1.706 & 3.742 & 0.103 & 0.041 & 0.170 & 0.903 & \textbf{0.965} &\textbf{0.983} \\ 
                                    & \acrshort{cafnet} (Ours)       & \checkmark &  \checkmark & \textbf{1.674} & \textbf{3.674} & \textbf{0.098} & \textbf{0.038} & \textbf{0.164} & \textbf{0.906} & 0.963 & \textbf{0.983} \\
                                    \hline
\multirow{6}{*}{70m}                & BTS \cite{bts} & \checkmark &            & 2.346 & 4.811& 0.119 & 0.047                                     & 0.188 & 0.872 & 0.952 & 0.979       \\ 
                                    & RC-PDA  \cite{rc-pda}   & \checkmark & \checkmark & 3.338 & 6.653 & 0.122  & 0.060 & 0.225 & 0.822 & 0.923 & 0.965  \\ 
                                    & RC-PDA-HG  \cite{rc-pda}  & \checkmark & \checkmark & 3.514 & 7.070 & 0.127  & 0.062 & 0.235 & 0.812 & 0.914 & 0.960      \\ 
                                    & DORN  \cite{dorn_radar}  & \checkmark & \checkmark & 2.170 & 4.532 & 0.105  & 0.055 & \textbf{0.170} & 0.896 & 0.960 & 0.982       \\ 
                                    & RadarNet    \cite{radarnet}  & \checkmark &  \checkmark & 2.073 & 4.591 & 0.105  & 0.043 & 0.181 & 0.896 & \textbf{0.962} & 0.981       \\ 
                                    & \acrshort{cafnet} (Ours) & \checkmark &  \checkmark &  \textbf{2.010} & \textbf{4.493} & \textbf{0.101} & \textbf{0.040} & 0.174 & \textbf{0.897} & 0.961  & \textbf{0.983} \\
                                    \hline
\multirow{10}{*}{80m}                & BTS \cite{bts}   & \checkmark &            & 2.467 & 5.125 & 0.120 & 0.048 & 0.191 & 0.869 & 0.951 & 0.979 \\ 
                                    & AdaBins \cite{adabins} & \checkmark &            & 3.541 & 5.885 & 0.197& 0.089 & 0.261 & 0.642 & 0.929 & 0.977       \\ 
                                    & P3Depth \cite{p3depth}  & \checkmark &            & 3.130 & 5.838 & 0.165& 0.065 & 0.222 & 0.804 & 0.934 & 0.974        \\ 
                                    & LapDepth \cite{lapdepth}   & \checkmark &            & 2.544 & 5.151 & 0.117 & 0.049 & 0.187 & 0.865 & 0.953 & 0.981       \\ 
                                    & RC-PDA    \cite{rc-pda}  & \checkmark & \checkmark & 3.721 & 7.632  &  0.126 & 0.063 & 0.238 & 0.813 & 0.914 & 0.960       \\ 
                                    & RC-PDA-HG  \cite{rc-pda}  & \checkmark & \checkmark & 3.664 & 7.775 & 0.138 & 0.064 & 0.245 & 0.806 & 0.909 & 0.957       \\ 
                                    & DORN  \cite{dorn_radar}  & \checkmark & \checkmark & 2.432 & 5.304 & 0.107 & 0.056 & 0.177 & 0.890 & 0.960 & \textbf{0.981}       \\ 
                                    & RCDPT$^{\dag}$ \cite{rcdpt}  & \checkmark & \checkmark & - & 5.165 & 0.095 &  - & - &0.901 & \textbf{0.961} & \textbf{0.981}      \\ 
                                    & RadarNet  \cite{radarnet}  & \checkmark & \checkmark & 2.179 & 4.899 & 0.106 & 0.044 & 0.184 & 0.894 & 0.959 & 0.980 \\ 
                                    & \acrshort{cafnet} (Ours)  & \checkmark & \checkmark & \textbf{2.109} & \textbf{4.765} & \textbf{0.101} & \textbf{0.040} & \textbf{0.176} & \textbf{0.895} & 0.959 & \textbf{0.981}\\
                                    \hline
\end{tabular}
}
\begin{tablenotes}
      \small
      \item $^{\dag}$ These results come from the paper that tests the model performance on a different test set, leading to the metrics being less comparable.
    \end{tablenotes}
     \vspace{-5mm}
    
\end{table*}

\subsection{Loss Functions}
Our model employs an end-to-end training strategy with several loss functions. For depth estimation, we use an L1 loss on the predictions $\hat{D}_{c}$ and $\hat{D}_{f}$ from both the coarse and final stages, respectively. The depth loss is given by:
\begin{equation}
\vspace{-1.3mm}
\begin{split}
    L_{Depth} = \frac{m}{|\Omega|}\sum_{x\in \Omega}|D_{acc}(x) - \hat{D}_{c}(x)|  \\
    + \frac{1}{|\Omega|}\sum_{x\in \Omega}|D_{acc}(x) - \hat{D}_{f}(x)|,
\end{split}
\end{equation}
where $m$ is a weighting factor, chosen as 0.5, and $\Omega$ represents the set of pixels where $D_{acc}$ is valid.

Additionally, an edge-aware smoothness loss is applied to the final depth map, $\hat{D}_{f}$, follows \cite{lin2020depth}, formulated as:
\begin{equation}
\vspace{-1.3mm}
    L_{Smooth} = |\nabla_{u} \hat{D}_{f}|e^{-|\nabla_{u}(I)|} + |\nabla_{v} \hat{D}_{f}|e^{-|\nabla_{v}(I)|},
\end{equation}
where $\nabla_{u}$ and $\nabla_{v}$ are the horizontal and vertical image gradients, respectively.

For radar confidence, we minimize a binary cross-entropy loss between the ground truth $C$ and the predicted map $\hat{C}$:
\begin{equation}
\vspace{-1mm}
\begin{split}
    L_{Conf} = -\frac{1}{|\Omega_{C}|}\sum_{x\in \Omega_{C}}\big(C(x)\log(\hat{C}(x)) \\
    + (1 - C(x))\log(1 - \hat{C}(x))\big),
\end{split}
\end{equation}
where $\Omega_{C}$ denotes the image domain.

The final loss $L$ is a weighted sum of the individual losses $L = L_{Depth} + L_{Conf} + \lambda L_{Smooth}$.
Here, $\lambda$ is a hyperparameter, chosen as $1e^{-3}$ based on empirical results.

%% file: chapter/experiments.tex
This section first describes the dataset and implementation setups. Then, we introduce the evaluation metrics and compare our method with existing approaches in qualitative and quantitative ways. Afterwards, we show the efficacy of the radar confidence map. Finally, we conducted ablation studies to underscore the robustness of the proposed techniques.

\subsection{Dataset and Implementation Details}
We utilize nuScenes \cite{nuscenes}, a leading-edge dataset for autonomous driving, to train and test our model's performance. This dataset is captured via multiple sensors, including cameras, radars, and lidar, mounted on a vehicle navigating through Boston and Singapore. Here, we utilize 700 scenes for training, 150 for validation, and 150 for testing.

During training, we follow \cite{radarnet} to accumulate 80 future and 80 past lidar frames by projecting the lidar point cloud at each frame to the current frame, thereby creating a denser depth map. Subsequently, we apply scaffolding \cite{scrffolding} to derive the interpolated depth map $D_{acc}$. During evaluation, we use the single-frame lidar depth map $D_{gt}$ provided by \cite{nuscenes}.

Our model is implemented using PyTorch \cite{Pytorch} and trained on a Nvidia\textsuperscript{\textregistered} Tesla A30 GPU. We adopt the Adam optimizer, starting with an initial learning rate of $1e^{-4}$ and applying a polynomial decay rate with a power of $p=0.9$. The training process spans 150 epochs with batches of 10. To mitigate overfitting, image augmentation techniques such as random flipping and adjustments to contrast, brightness, and color
are employed. Additionally, random cropping to $352 \times 704$ pixels is performed during training to further enhance the model's robustness.

\subsection{Evaluation Metrics}
Table. \ref{table:metrics} defines the evaluation metrics, where the predicted and ground truth depth maps are represented by $\hat{d}$ and $d_{gt}$. $\Omega$ denotes the set of pixels where $d_{gt}$ is valid.
\begin{table}[ht]
\vspace{-2mm}
\caption{Metrics definition for depth estimation task.}
\vspace{-1mm}
\centering
\begin{tabular}{l||cccc}
\hline
 & Definition\\ \hline
MAE  &  $\frac{1}{|\Omega|}\sum_{x\in \Omega} |\hat{d}(x)-d_{gt}(x)|$      \\ 
RMSE    &   $ (\frac{1}{|\Omega|}\sum_{x\in \Omega} |\hat{d}(x)-d_{gt}(x)|^{2})^{1/2} $  \\
AbsRel       & $\frac{1}{|\Omega|}\sum_{x\in \Omega} |\hat{d}(x)-d_{gt}(x)|/d_{gt}(x)$ \\
log10  & $\frac{1}{|\Omega|}\sum_{x\in \Omega} |\log_{10}\hat{d}(x)-\log_{10}d_{gt}(x)|$ \\
RMSElog & $\sqrt{\frac{1}{|\Omega|}\sum_{x\in \Omega} ||\log_{10}\hat{d}(x)-\log_{10}d_{gt}(x)||^{2}}$\\
$\delta_{n}$ threshold   &   $\delta_{n}=|\{\hat{d}(x): max(\frac{\hat{d}(x)}{d_{gt}(x)}, \frac{d_{gt}(x)}{\hat{d}(x)})< 1.25^{n}\}|/|\Omega|$        \\
 \hline
\end{tabular}
\label{table:metrics}
\vspace{-3.5mm}
\end{table}
\begin{figure*}[ht]
\centering
\includegraphics[width=0.98\textwidth]{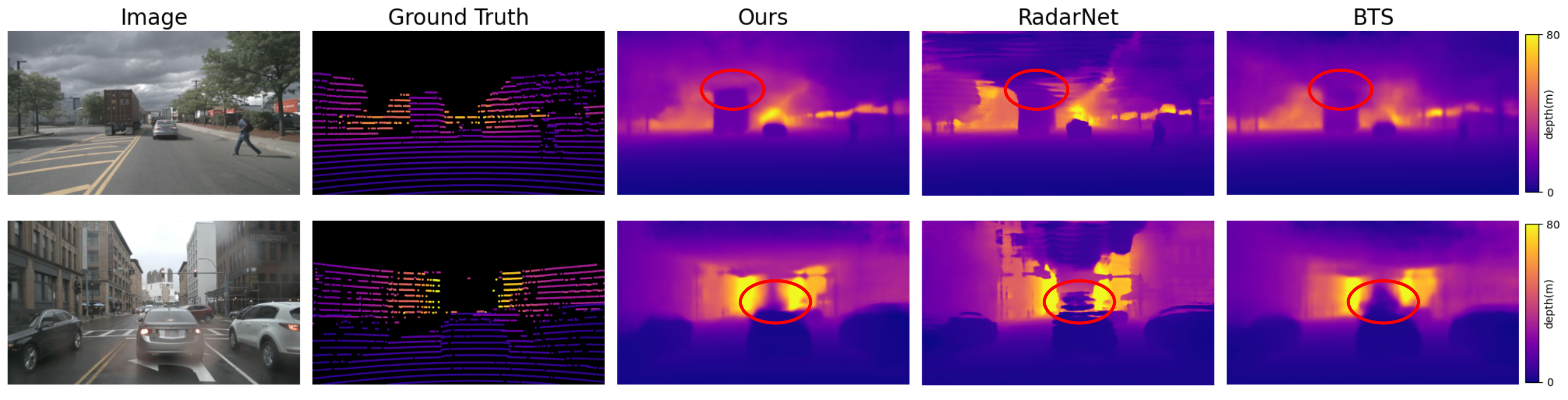}

\vspace{-3mm}
   \caption{Qualitative comparison on nuScenes test set. Column 1 shows the RGB image; column 2 plots the ground truth depth map. We compare our result with the RadarNet and our baseline BTS at 80 meters depth range.}
\label{fig:qualitative}
\vspace{-4mm}
\end{figure*}
\subsection{Quantitative Results}
We benchmark our method against existing image-only techniques \cite{adabins,p3depth,bts,lapdepth} and radar-camera fusion strategies \cite{dorn_radar,rc-pda,radarnet,rcdpt} on the official nuScenes test set, utilizing metrics in TABLE \ref{table:metrics}. 
We conduct comparisons at 50, 70, and 80-meter intervals, with the results presented in TABLE \ref{tab:exp}. Our model surpasses image-only depth estimators across all metrics. Specifically, comparing to our baseline \cite{bts}, our approach exhibits an MAE reduction of $13.6\%$, $14.3\%$, and $14.5\%$, and RMSE decrease of $5.4\%$, $6.6\%$, and $7.0\%$ at the respective distances. These improvements underscore the advantage of integrating radar data, which reliably captures long-range details that are often not visible in RGB imagery.

Following \cite{radarnet}, our framework utilizes a single radar scan to ensure feasibility in real-world applications. In contrast, techniques such as those presented in \cite{lin2020depth,dorn_radar,rc-pda} leverage multiple radar sweeps to augment the dataset.
Despite the limited radar data, our proposed technique demonstrates superior performance in nearly all evaluated metrics. Compared to the leading model, RadarNet \cite{radarnet}, we achieve an improvement of $1.9\%$, $3.0\%$, and $3.2\%$ in MAE for the stated ranges, further establishing the efficacy of our approach.

\subsection{Qualitative Results}

Fig. \ref{fig:qualitative} compares the performance of our method with the baseline \cite{bts} and the leading fusion model \cite{radarnet} across distances up to 80 meters under two scenarios. In the first scenario, our model precisely identifies the truck within the indicated regions (highlighted with red circles), unlike the other methods, which merge the truck's top with the sky. Moreover, BTS fails to recognize the pedestrian, who is nearly visible on its depth map. 
The second row illustrates that under rainy conditions, the resulting blurred RGB image hinders the other models from delivering an accurate prediction. For instance, the upper part of the central vehicle can hardly be detected.
In contrast, our method robustly differentiates and correctly measures the depth of each object. Additionally, depth maps produced by RadarNet often exhibit discontinuities, which can constrain the 3D reconstruction process. Our method, however, generates accurate depth maps that clearly distinguish individual objects.

\subsection{Efficacy of Radar Confidence}
\label{subsec:radarconf_exp}
\begin{figure}
    \begin{subfigure}[b]{.49\columnwidth}
        \includegraphics[width=\linewidth]{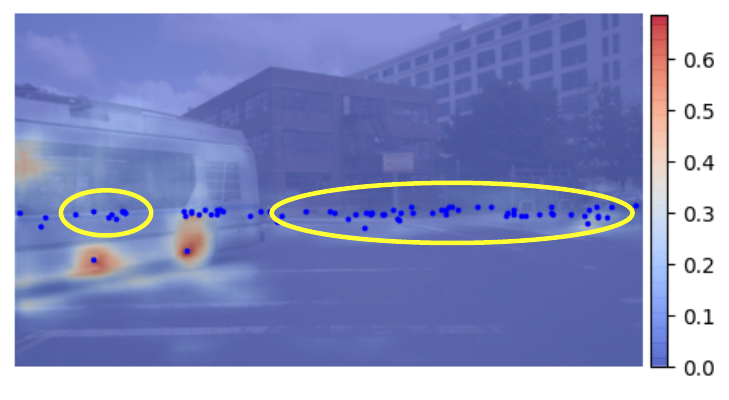}
    \end{subfigure}
    \hfill
    \begin{subfigure}[b]{0.49\columnwidth}
        \includegraphics[width=\linewidth]{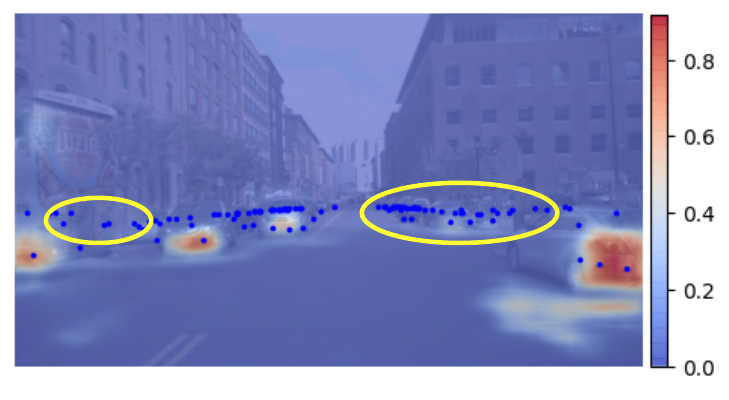}
    \end{subfigure}

\caption{Radar confidence map on top of the RGB image. The blue dots visualize the radar projected points. The confidence map indicates the possible projection surface of the radar points.}
\label{fig:rad_conf}
\vspace{-4mm}
\end{figure}

Fig. \ref{fig:rad_conf} showcases the radar confidence map with radar points marked in blue. The maps indicate the probable projection region for each radar point, aiming to mitigate the effects of noisy and inaccurate radar detections.
Areas highlighted in red are identified as highly probable projection zones. Notably, as highlighted in yellow circles, a majority of the radar points are encompassed by regions with markedly low confidence scores. This observation indicates that these points are predominantly noise, which could lead to misleading during training. Involving this confidence map in the fusion process reduces the influence of such noisy data, thereby preserving and emphasizing valuable information.

\begin{figure}
    \begin{subfigure}[b]{.49\columnwidth}
        \includegraphics[width=\linewidth]{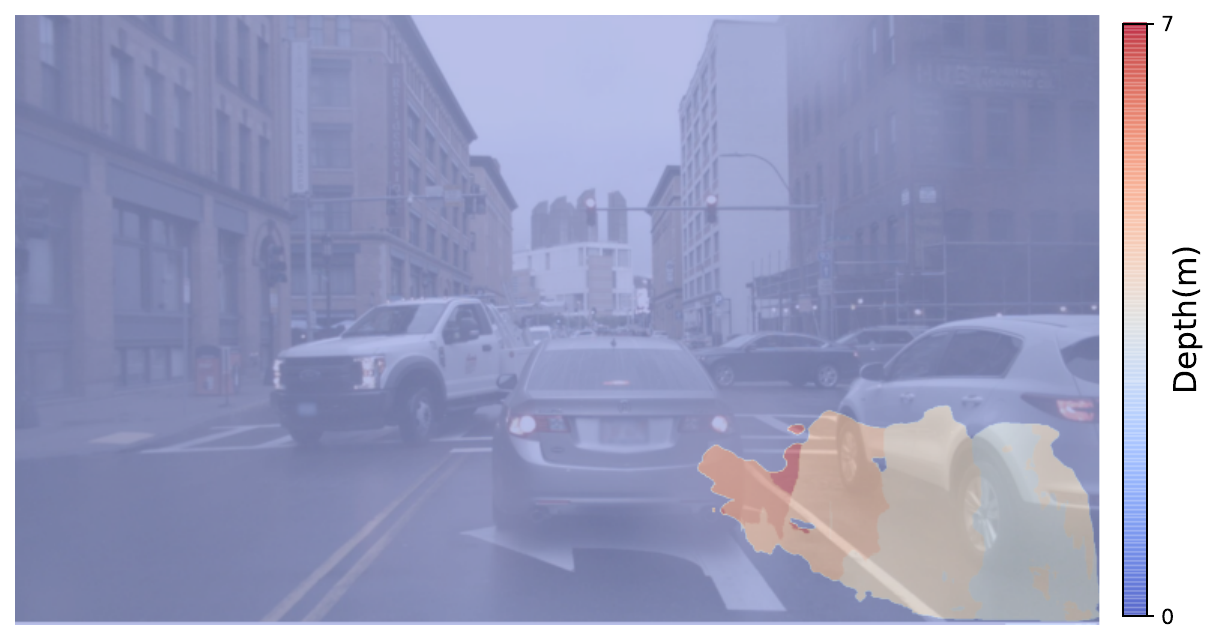}
        \caption{Quasi-dense map estimated from the first stage of RadarNet \cite{radarnet}.}
        \label{fig:radnet_quasi}
    \end{subfigure}
    \hfill
    \begin{subfigure}[b]{0.49\columnwidth}
        \includegraphics[width=\linewidth]{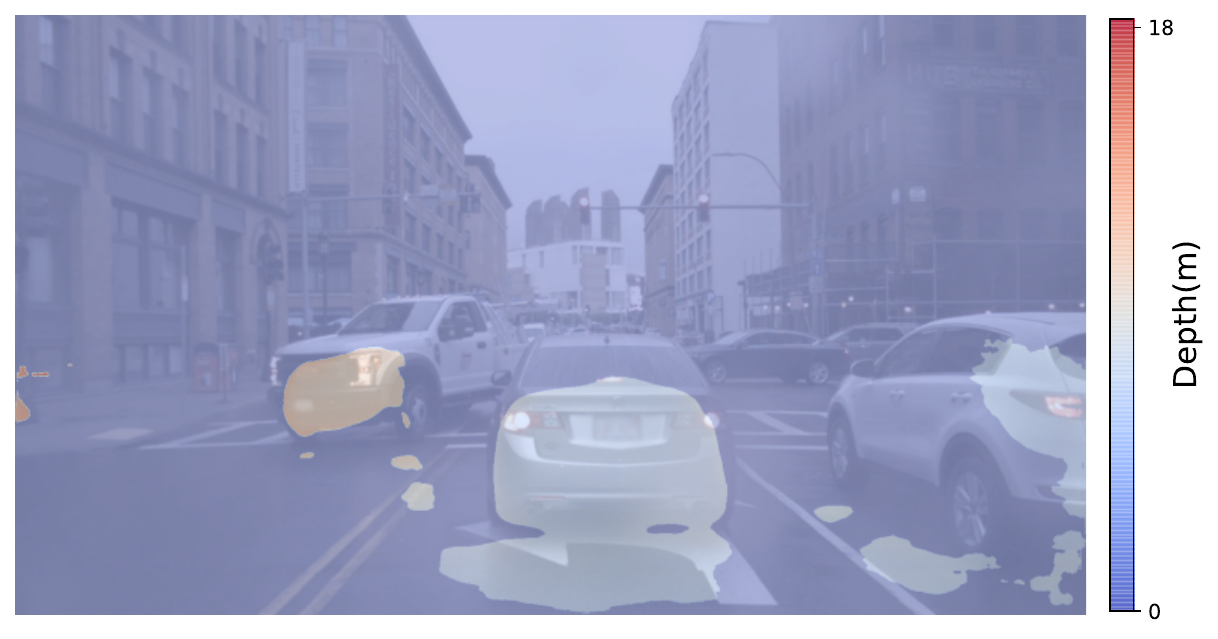}
        \caption{Generated refined depth after the refinement module.}
        \label{fig:refine_depth
        }
    \end{subfigure}
\vspace{-3.3mm}

\caption{Qualitative comparison between the quasi-dense map~\cite{radarnet} and our refined depth map. }
\label{fig:refinedepth_quasi_comp}
\vspace{-7mm}
\end{figure}

Subsequently, the refinement module processes $\hat{C}$ and $\hat{D}_{c}$ to produce a refined depth map. We then evaluate our refined depth map against the quasi-dense map generated by RadarNet \cite{radarnet}. In RadarNet, the quasi-dense maps generated during the initial stage serve as the radar input for the subsequent stage. As depicted in Fig. \ref{fig:refinedepth_quasi_comp}, our refined depth map showcases significant depth details at locations corresponding to radar points with a confidence score exceeding 0.4. Conversely, the quasi-dense map from \cite{radarnet} includes data that lacks coherence and reliability, rendering it unsuitable as a dependable input for the final depth estimation process.

\begin{table}[ht]
\vspace{-2mm}
\caption{Ablation study about the radar confidence.}
\centering
\begin{tabular}{l||cccc}
\hline
 & MAE $\downarrow$ & RMSE $\downarrow$ & AbsRel $\downarrow$ & $\delta_{1}$ $\uparrow$\\ \hline
GT from \cite{radarnet}   & 2.192  &  4.831  &  0.106 & 0.889     \\
GT from \cite{rc-pda}   & 2.196  &  4.838  &  0.106 & 0.890     \\
w/o RM   & 2.189  &  4.844  & 0.105  & 0.890      \\
\acrshort{cagf}$\rightarrow$ Add
& 2.201   &   4.855      & 0.108 &  0.891 \\
\acrshort{cagf}$\rightarrow$ Concat
& 2.198   &   4.849      & 0.107 &  0.890 \\
\acrshort{cagf}$\rightarrow$ GF  & 2.194   &   4.846      & 0.107 &  0.889 \\
w/o confidence  & 2.208 & 4.878 & 0.109 & 0.890 \\
Ours & \textbf{2.109} & \textbf{4.765} & \textbf{0.101} & \textbf{0.895} \\ 

 \hline
\end{tabular}
\label{table:ablation_radar_conf}
\vspace{-5.5mm}
\end{table}

\subsection{Ablation Study}
We conduct ablation studies to ascertain the impact of the proposed components. The results are detailed as follows:

\noindent \textbf{Effectiveness of radar confidence:} TABLE \ref{table:ablation_radar_conf} showcases the experimental outcomes related to radar confidence.
First, we benchmark our ground truth (GT) generation approach against the method proposed in \cite{radarnet} and \cite{rc-pda}. Compared to \cite{radarnet}, which defines a selective region of $288\times 900$, our methodology yields a $3.8\%$ improvement in MAE. 

Secondly, in the Refinement Module (RM), $\hat{C}$ is employed to generate the confidence-refined depth, concatenated with $R$, and progressed to the second stage. We evaluate this approach against a baseline that directly concatenates $\hat{D}_{c}$, $\hat{C}$, and $R$, rather than refinement. The absence of RM leads to an overload of detailed information alongside sparse radar data, potentially misleading the model's feature extraction.

Most importantly, we show the effectiveness of the \acrshort{cagf} by comparing it with the Gated Fusion (GF)\cite{radarnet} and the element-wise addition and concatenation. Our \acrshort{cagf} is able to distinguish the relevance of individual pixels within radar features during the fusion process, enhancing MAE by $3.9\%$.

Finally, to underscore the significance of the confidence map, we designed an additional experiment in which the first stage only predicts a coarse depth map $\hat{D}_{c}$. Consequently, within the RM, $\hat{D}_{c}$ is concatenated with $R$ and processed by the second encoder. In the decoding phase, the GF mechanism is employed to fuse the radar and image features. This setup aims to show the vital role of the confidence map in enhancing the accuracy and reliability of depth estimation.

\begin{table}[ht]
\vspace{-2mm}
\caption{Ablation study about the SCM.}
\vspace{-1mm}
\centering
\begin{tabular}{l||cccc}
\hline
 & MAE $\downarrow$ & RMSE $\downarrow$ & AbsRel $\downarrow$ & $\delta_{1}$ $\uparrow$\\ \hline
w/o SCM  & 2.138 & 4.780 & 0.105 & 0.892 \\
Ours (final) & \textbf{2.109} & \textbf{4.765} & \textbf{0.101} & \textbf{0.895} \\ 
 \hline
\end{tabular}
\label{table:ablation_scm}
\vspace{-3.5mm}
\end{table}

\noindent \textbf{Effectiveness of SCM: }
Table \ref{table:ablation_scm} presents the outcomes demonstrating that the integration of the SCM leads to a $1.4\%$ enhancement in MAE. This improvement underscores SCM's efficacy in extracting useful features from sparse data.

%% file: chapter/conclusion.tex
Depth estimation task using radar and camera data presents considerable challenges, primarily due to the inherent sparsity and ambiguity in radar data. In response to these challenges, we introduced \acrshort{cafnet} that operates in two stages. The first stage generates a radar confidence map and a preliminary depth map. 
In the second stage, we implement an innovative fusion technique that leverages the confidence scores to enhance the feature fusion process. This method discriminates between the radar pixels based on their informational value, without incorporating additional misleading data, thereby significantly improving the accuracy of the depth maps. Our evaluations of the nuScenes dataset demonstrate the superiority of our approach over existing methodologies. Through our research, we provide a robust solution of radar-camera depth estimation, which improves the MAE by $3.2\%$. Moving forward, we aim to enhance the utilization of radar data by incorporating 3D geometric information more comprehensively.

%% file: chapter/acknow.tex
Research leading to these results 
has received funding from the EU ECSEL Joint Undertaking under grant agreement n° 101007326 (project AI4CSM) and from the partner national funding authorities the German Ministry of Education and Research (BMBF).